\documentclass[letterpaper, 10 pt, conference]{ieeeconf}  

\IEEEoverridecommandlockouts                              

\usepackage{graphicx} 
\usepackage{amsmath} 
\usepackage{amssymb}  

\usepackage{amsthm}
\usepackage{mathtools}
\usepackage{soul}
\usepackage{dsfont}
\usepackage{gensymb}
\usepackage{booktabs}
\usepackage{makecell,array}
\usepackage{accents}
\usepackage[table]{xcolor}
\usepackage{booktabs}
\usepackage{colortbl}
\usepackage{multirow}
\usepackage{siunitx}

\renewcommand{\arraystretch}{1.8}

\usepackage{bm}
\theoremstyle{definition}

\newtheorem{assumption}{Assumption}

\usepackage[caption=false]{subfig}

\usepackage{caption}
\DeclareCaptionType{equ}[][]

\usepackage{hyperref}
\usepackage{algorithm}  
\usepackage{algpseudocode}[noend]

\usepackage{todonotes}
\usepackage{cite}

\newcommand{\traj}{\zeta}
\newcommand{\inpsig}{\mathbf{u}}
\newcommand{\nat}{\mathbb{N}}

\graphicspath{{figures/}} 

\title{\LARGE \bf
GPU-Accelerated Barrier-Rate Guided MPPI Control for Tractor-Trailer Systems

}

\author{Keyvan Majd$^{1}$, Hardik Parwana$^{2}$, Bardh Hoxha$^{1}$, Steven Hong$^{1}$, Hideki Okamoto$^{1}$, Georgios Fainekos$^{1}$\\ 
\thanks{$^{1}$Toyota Motor North America, Research \& Development, Ann Arbor, MI 48105, USA. {\tt\small <first\_name.last\_name>@toyota.com}}%
\thanks{$^{2}$The work was performed during an internship at Toyota Motor North America, Research \& Development}%
\thanks{© 2025 IEEE. Personal use of this material is permitted. Permission from IEEE must be obtained for all other uses, in any current or future media, including reprinting/republishing this material for advertising or promotional purposes, creating new collective works, for resale or redistribution to servers or lists, or reuse of any copyrighted component of this work in other works.}
}

\newcommand{\reals}{\mathbb{R}}

\newcommand{\X}{\mathcal{X}}
\newcommand{\U}{\mathcal{U}}

\newcommand{\K}{\mathcal{K}}
\newcommand{\Obst}{\mathcal{O}}    
\newcommand{\classK}{class-$\K$ }

\newcommand{\classKinf}{\mbox{class-$\K_\infty$}}

\newcommand{\posX}{p^x}                      
\newcommand{\posY}{p^y}                      
\newcommand{\speed}{v}                     
\newcommand{\accel}{a}                     
\newcommand{\heading}{\theta}     
\newcommand{\headingTractor}{\heading^{1}}     
\newcommand{\headingTrailer}{\heading^{2}}     
\newcommand{\steerAngle}{\delta}           
\newcommand{\jerk}{j}                      
\newcommand{\steerRate}{\omega}            
\newcommand{\wheelbase}{\ell_1}              
\newcommand{\hitchLen}{\ell_h}               
\newcommand{\trailerLen}{\ell_2}             
\newcommand{\Ts}{T_s}                        
\newcommand{\contourProgress}{\psi}          
\newcommand{\contourProgressRate}{\nu}          
\newcommand{\refPath}{\zeta}          
\newcommand{\refPosX}{p^{x,r}}                      
\newcommand{\refPosY}{p^{y,r}}                      
\newcommand{\refHeadingTractor}{\heading^{1,r}}     
\newcommand{\refHeadingTrailer}{\heading^{2,r}}     
\newcommand{\refPathLen}{L}

\newcommand{\obsPosX}{p^x_o}           
\newcommand{\obsPosY}{p^y_o}           
\newcommand{\obsPosXPrime}{p^{x'}_o}
\newcommand{\obsPosYPrime}{p^{y'}_o}
\newcommand{\obsSemiAxisX}{m_x}        
\newcommand{\obsSemiAxisY}{m_y}        
\newcommand{\obsHeading}{\theta_o}     
\newcommand{\barrier}{h}           

\newcommand{\velScale}{\kappa}                   
\newcommand{\hitchThreshMod}{\hitchThresh - \velScale\,\speed}  

\newcommand{\hitchDelta}{\Delta\theta}           
\newcommand{\hitchThresh}{\Delta\theta_{\max}}   

\newcommand{\contourErr}{e^c}                
\newcommand{\contourErrWeight}{Q_c}          
\newcommand{\lagErr}{e^l}                    
\newcommand{\lagErrWeight}{Q_l}              
\newcommand{\headingErr}{e^\heading}                    
\newcommand{\headingErrWeight}{Q_\heading}              
\newcommand{\controlEffortWeight}{R}              
\newcommand{\contourProgressRateWeight}{\mu}              
\newcommand{\stageCost}{r}

\newcommand{\sample}{s}
\newcommand{\numSamples}{S}

\newcommand{\augInput}{z}                     
\newcommand{\augInputDes}{z_{\mathrm{des}}}   
\newcommand{\augInputLow}{\underline{z}}      
\newcommand{\augInputHigh}{\overline{z}}      
\newcommand{\Wmat}{W}                         

\begin{document}

\maketitle
\thispagestyle{empty}
\pagestyle{empty}



\begin{abstract}

Articulated vehicles such as tractor-trailers, yard trucks, and similar platforms must often reverse and maneuver in cluttered spaces where pedestrians are present. 
We present how Barrier-Rate guided Model Predictive Path Integral (BR-MPPI) control can solve navigation in such challenging environments.
BR-MPPI embeds Control Barrier Function (CBF) constraints directly into the path-integral update. 
By steering the importance-sampling distribution toward collision-free, dynamically feasible trajectories, BR-MPPI enhances the exploration strength of MPPI and improves robustness of resulting trajectories.
The method is evaluated in the high-fidelity CarMaker simulator on a 12 [m] tractor-trailer tasked with reverse and forward parking in a parking lot. 
BR-MPPI computes control inputs in above 100 [Hz] on a single GPU (for scenarios with eight obstacles) and maintains better parking clearance than a standard MPPI baseline and an MPPI with collision cost baseline.

\end{abstract}


\section{Introduction}


Advanced Driver Assistance Systems (ADAS) have enhanced passenger vehicle safety and driver convenience. 
While ADAS technologies such as adaptive cruise control, emergency braking, dynamic obstacle avoidance, and parking assist have become standard in many consumer vehicles, 
their implementation in tractor-trailer systems and recreational towing remains underdeveloped. 
The unique challenges posed by heavy-duty and towing scenarios necessitate specialized solutions.
These solutions must account for multi-body vehicle dynamics, uncertain weight distribution, and highly unstable reverse motion dynamics, which are not typically addressed by existing ADAS frameworks. 
Research into this domain offers the potential to significantly enhance safety in recreational towing and mitigate driver fatigue in commercial towing operations.

In this work, a control architecture is presented that addresses two key technical challenges: automated parking assist for tractor-trailer systems and dynamic obstacle avoidance. 
In recreational towing, suitable trailer parking locations are frequently not accessible through straightforward, easily computable maneuvers. 
Multi-point turning maneuvers may be required to access confined parking spaces, such as those encountered at camping grounds. 
In dynamic obstacle avoidance, an emergency collision avoidance system must consider not only the movement of the obstacle but also the movement of the trailer. 
The proposed framework is capable of addressing such technical challenges uniformly. 

We build upon Barrier-Rate guided Model Predictive path Integral (BR-MPPI) control \cite{parwana2025brmppi} to develop a real-time control architecture that can address Tractor-Trailer (TT) navigation challenges.
BR-MPPI integrates Model Predictive Path Integral (MPPI) \cite{williams2018information} control with Control Barrier Functions (CBF) \cite{ames2016control} for efficient and safe configuration space exploration.
At its core, MPPI iteratively simulates random future scenarios and computes an optimal trajectory by employing a cost function that assesses the fitness of these simulations. 
In contrast to traditional Model Predictive Control (MPC) methodologies \cite{lam2010model,VallinderEtAl2024itsc,liniger2015optimization} that necessitate explicit gradient calculations, MPPI employs probabilistic sampling and weighted averaging, thereby handling nonlinear, uncertain, and/or discontinuous dynamics \cite{parwana2025brmppi}. 
On the other hand, CBFs facilitate the mathematical encoding of the safe operational domain of an autonomous system among static and dynamic obstacles. 

A rudimentary integration of MPPI with CBF would involve utilizing CBF as a filter on the trajectories returned by MPPI. 
Nevertheless, this approach exhibits limitations manifested as either excessive conservatism within complex environments or the formulation of infeasible optimization problems.
This shortcoming primarily arises from two factors. 
First, the reliance of MPPI on soft constraints through the cost function hinders the ability to generate collision-free trajectories within constrained search spaces. 
Secondly, the myopic nature of CBF, which prioritizes immediate obstacle avoidance, can result in infeasible solutions when safety constraints cannot be satisfied.

BR-MPPI addresses these issues by incorporating CBF constraints in the sampling process itself.
Specifically, during sampling of control actions and parameters within MPPI, samples that fail to satisfy safety constraints are projected onto the safe subspace.
This approach enhances the exploration of narrow spaces, which is a capability lacking in naive free space sampling, e.g., standard MPPI \cite{williams2018information}.
This is a core algorithmic capability required for planning reverse maneuvers in TT systems.

Our contributions can be summarized as follows:
\begin{enumerate}
    \item we formulate the BR-MPPI cost terms that are necessary to handle TT navigation applications and we show that the standard MPPI cannot solve similar problem with soft obstacle avoidance cost,
    \item we modify the BR-MPPI's projection operator as a soft penalty, with penalty weights tuned using temporal-logic-robustness guided falsification,
    \item we implement the framework in JAX enabling real-time execution on GPU platforms, and
    \item we evaluate the framework by simulating a pickup truck towing and parking a boat within a high-fidelity simulator (CarMaker).
\end{enumerate}


The first contribution is fundamental in solving the reverse maneuvers in tractor-trailer (TT) systems.
As we elaborate in Section \ref{sec:method}, some constraints are captured in the cost function, while others, e.g., dynamic obstacles and hitch angle constraints, must be imposed through CBF constraitns.
Finally, an important benefit of taking an MPPI approach is that MPPI approaches can handle data-driven enhanced models without any changes to the framework, e.g., \cite{LongEtAl2024arxivContinuum}.

\begin{figure*}[t]
    \centering
    \includegraphics[width=1\linewidth]{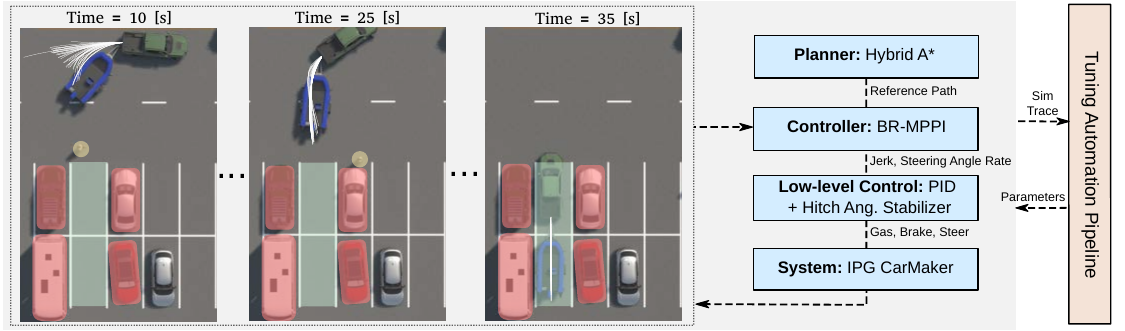}
    \caption{\textbf{System architecture:} A high-level planner produces a rough feasible path with attempt to satisfy motion constraints, but it is not guaranteed to. Then, BR-MPPI uses the reference path to compute a dynamically feasible and safe trajectory for a short horizon while avoiding obstacles (static red cars and dynamic yellow human). Finally, the trajectory computed by BR-MPPI is translated to accelerator, brake, and steering commands for the vehicle. The control loop repeats until the trailer has reached the target location (green parking spot).
    }
    \label{fig:overview}
\end{figure*}

\section{Related Research}







Integrating Control Barrier Functions (CBFs) with MPPI is a promising area for improving safety. 
Unlike BR-MPPI's direct CBF embedding and sample projection, other methods explore alternative integrations. 
Guaranteed-Safe MPPI \cite{RabieeH2024arxivMPPIcbf} synthesizes a safe closed-loop system by composing the original plant with a closed-form CBF.
The resulting closed-loop system can be used for safe sampling through MPPI.
However, the composed system is too conservative and shrinks exploration when constraints are tight.
Shield-MPPI \cite{YinDFT2023ral} runs a sampler that accounts for CBF constraint violation costs. 
Then, following the standard CBF filter practice, it overwrites unsafe inputs with a quadratic CBF shield. 
BR-MPPI achieves better space exploration since within the MPPI loop, the sampled trajectories are projected onto the safe space defined by the CBF.
Methods like \cite{TaoEtAl2022acc} and \cite{WangCT2025arxivMPPIdbas} adapt the sampling process so that they can sample with higher probability from the safe set defined by the CBF.
Finally, \cite{BorquezRCB2025arxivMPPI} uses Hamilton-Jacobi reachability in DualGuard-MPPI for provably safe samples, but the approach is computationally expensive. 
In brief, when compared to all the aforementioned methods, BR-MPPI explores the obstacle free space more effectively because it also samples CBF parameters within the MPPI loop, which reduces conservatism. 

On the application front, motion planning and control problems for articulated vehicles have been studied for a long time.
The forward path tracking problem for TT systems still remains an active research area \cite{LiuYZZ2025its,SelimBJ2024itsc,VallinderEtAl2024itsc,KangEtAl2024itsc}.
However, our work specifically focuses on the reversing problem for articulated vehicles.
The reversing problem poses significant challenges due to inherent instability risks which can lead to jackknifing.
One of the first applications of Neural Network based control to the TT reversing problem appears in \cite{NguyenW1989}.
Later, \cite{AltafiniSJ2002hscc} presented a hybrid control architecture for the reversing problem.
A control safety governor approach is presented in \cite{HejaseEtAl2018itsc} by deriving maneuverability conditions for TT systems that are hitched off-axle.
The works most closely related to ours appear in \cite{WangEtAl2022ccta,GaoJLYS2025cep} where Nonlinear MPC (NMPC) approaches are presented.
In particular, \cite{GaoJLYS2025cep} combines NMPC with CBF.
Even though the performance and functionality of the NMPC approaches are similar to our BR-MPPI approach, MPPI approaches have some distinct advantages.
Namely, MPPI approaches can incorporate data-enabled models \cite{LongEtAl2024arxivContinuum} operating in highly uncertain environments \cite{williams2018information}.
Our work sets the foundations for addressing the TT-reversing problem using MPPI based approaches.

Finally, solving navigation problems for TT systems in complex environments requires a path planner \cite{LeuWTC2022iros,HellanderBA2024itsc,MaEtAl2024itsc,ZhaoEtAl2024its} in addition to path tracking.
Our proposed BR-MPPI approach can take advantage of any improvements on path generation. 
In addition, due to the local search capabilities of MPPI approaches even a rough approximate initial path can be a good candidate for solving the navigation problem up to the required accuracy.

\section{Preliminaries}
\subsection{Notation}

The set of real numbers, and positive real numbers are denoted as $\reals$, and $\reals^+$ respectively. 
The set of natural numbers (with zero) is denoted by $\nat$ and an interval in $\nat$ is denoted by $[a,b]_{\nat}$ or as a collection $\{k\}^{b}_{k=a}$.
For brevity, when we write ``for all $\{k\}^{b}_{k=a}$" we mean ``for all $k \in [a,b]_\nat$".

The time derivative of $x$ is denoted by $\dot x$. Given $x\in \reals^n$, $||x||_Q\coloneqq\sqrt{x^T Q x}$ is called the weighted $Q$ norm for a positive definite matrix $Q$. 
A continuous function $\alpha: [0,a)\rightarrow [0,\infty)$ is a \classK function if it is strictly increasing and $\alpha(0)=0$. 
Furthermore, if $\lim_{r \rightarrow \infty} \alpha(r) = \infty$, then it is called \classKinf. 
We denote a sequence $x_1, x_2,\cdots,x_N$ or tuple of variables $(x_1, x_2,\cdots,x_N)$ by $(x_i)^N_{i=1}$.
Finally, the Lie derivative of a differentiable \(b\colon\reals^n\to\reals\) along a vector field \(f\colon\reals^n\to\reals^n\) is denoted
$
L_f b(x) = \nabla b(x)\,\cdot\,f(x).
$
\subsection{Control Barrier Functions}
Consider the discrete-time control-affine system
\begin{equation}\label{eq:affine_dynamics}
  x_{k+1} = x_k + f(x_k) + g(x_k)\,u_k,
\end{equation}
where states $x_k\in\mathbb{R}^n$, control inputs $u_k\in \U \subseteq \mathbb{R}^m$, and the functions
$f:\mathbb{R}^n\to\mathbb{R}^n$ and $g:\mathbb{R}^n\to\mathbb{R}^{n\times m}$
are continuous. 
We assume that (\ref{eq:affine_dynamics}) is the result of zero-order hold on input $u_k$ and with constant sampling period $\Ts>0$ on a continuous-time system $\dot x = f_c(x)+g_c(x)u$ so that
$f(x)=\Ts\,f_c(x)$ and $g(x)=\Ts\,g_c(x)$.
For brevity in the presentation, we set $\Ts=1$ in~\eqref{eq:affine_dynamics}, but typically we need to consider stability and accuracy issues in order to determine an appropriate $\Ts$ \cite{VallinderEtAl2024itsc}.

A safe set $\mathcal{X}_s\subseteq\mathbb{R}^n$ can be defined as zero-superlevel set of
differentiable constraint function $\barrier:\mathbb{R}^n\to\mathbb{R}$ as
$\X_s \triangleq \{ x \in \X : \barrier(x) \geq 0 \}$.
A \emph{discrete-time Control Barrier Function} (CBF) is a candidate function $\barrier$
that enforces forward invariance of $\X_s$ under~\eqref{eq:affine_dynamics}.
For a class-$\mathcal{K}_\infty$ function $\beta$, a sufficient
\emph{discrete CBF condition} is given by~\cite{zeng2021safety,ames2016control}:
\begin{align}\label{eq:barier-const-beta}
    \sup_{u\in\U}\bigl[\barrier(x_{k+1}) - \barrier(x_k)\bigr]\ge \beta\bigl(\barrier(x_k)\bigr),
  \quad\forall x_k\in\X_s,
\end{align}
where $\U$ is the admissible input set. 
Assuming a linear class-$\mathcal{K}$ function $\beta(s)=\alpha s$, $\alpha\in(0,1]$, changes \eqref{eq:barier-const-beta} to
$
  \barrier(x_{k+1}) - \barrier(x_k) \ge \alpha\,\barrier(x_k).
$
Substituting~\eqref{eq:affine_dynamics} in \eqref{eq:barier-const-beta}, performing a first-order Taylor expansion
of $\barrier(x_{k+1})$ around $x_k$ and neglecting higher-order terms leads to the affine constraint:
\begin{align}\label{eq:barrier_constraint_prelim}
    L_f \barrier(x_k) + L_g \barrier(x_k)\,u_k \ge \alpha\,\barrier(x_k),
\end{align}
This linear constraint can be incorporated into optimization-based controllers, most commonly as a linear constraint in a quadratic program (often called a CBF‐QP \cite{ames2016control, zeng2021safety}), to ensure that each control input satisfies the barrier condition $\barrier$ and thus maintains safety.
\subsection{Model Predictive Path Integral (MPPI) Control}\label{sec:prelim_mppi}
MPPI \cite{williams2018information} is a recent sampling based Model Predictive Control (MPC) method.  
It approximates the solution of the stochastic optimal control problem:
\begin{align}\label{eq:mppi_cost_function}
        \min_{(u_\tau)^{k+H}_{\tau=k}}\quad &  \mathbb{E} \Biggl[  \sum_{\tau=k}^{k+H} \stageCost(x_\tau,u'_\tau) + \left( \frac{\lambda}{2} u_\tau^T \Sigma_w^{-1} u_\tau \right)  \Biggr], \\
   \textrm{s.t.} \quad & (\ref{eq:affine_dynamics}) \text{ and } w_\tau \sim \mathcal{N}(0,\Sigma_w), \nonumber
\end{align}
where \(\stageCost:\mathbb{R}^n\times\mathbb{R}^m\to\mathbb{R}^+\) is a user-defined stage cost (e.g., to penalize deviation from a target state or to enforce collision avoidance), 
$H$ is the prediction horizon, 
$u_{\tau}\in\mathbb{R}^m$ is the mean control input at time $\tau$, 
$w_\tau\sim\mathcal{N}(0,\Sigma_w)\subset\mathbb{R}^m$ is a zero-mean Gaussian noise with  covariance $\Sigma_w\in\mathbb{R}^{m\times m}$, 
and $u'_{\tau} = u_{\tau}+w_{\tau}$ denotes a disturbed control input.

At time \(k\), 
the MPPI algorithm
starts from a nominal control input sequence $(u_{\mathrm{nom},\tau})^{k+H}_{\tau=k}$ and 
draws \(\numSamples\) independent noise rollouts \(\tilde{w}^\sample_\tau\sim\mathcal{N}(0,\Sigma_w)\) for \(\tau=k,\dots,k+H\) that forms perturbed controls \(\tilde{u}^\sample_\tau=u_{\mathrm{nom},\tau}+\tilde{w}^\sample_\tau\) for $\{\sample\}^\numSamples_{\sample=1}$. 
It then propagates the perturbed controls through the system dynamics \eqref{eq:affine_dynamics} to obtain the trajectory $\traj_\sample=[x^\sample_k, \cdots,x^\sample_{k+H}]$.
Each trajectory $\traj_\sample$ accumulates cost
$
  J_\sample=\sum_{\tau=k}^{k+H}\stageCost(x^\sample_\tau,\tilde{u}^\sample_\tau),
$
from which the algorithm computes importance weights
\[
  \gamma_\sample = \exp\!\Bigl(-\tfrac{J_\sample-\min_j J_j}{\lambda}\Bigr),
  \qquad
  \bar\gamma_\sample = \frac{\gamma_\sample}{\sum_{j=1}^\numSamples\gamma_j},
\]
with “\textit{inverse temperature}” $\lambda>0$.  The nominal sequence is updated by
\begin{align}\label{eq:MPPI_update_rule}
    u_{\mathrm{nom},\tau} \;\leftarrow\;\sum_{\sample=1}^\numSamples \bar\gamma_\sample\,\tilde{u}^\sample_\tau,\quad \tau=k,\dots,k+H,
\end{align}
then \(u_k=u_{\mathrm{nom},k}\) is applied to the system. Next, the last input is shifted, and the process repeats in a receding-horizon fashion.

\section{Problem Formulation}

In this paper, we study the problem of tractor-trailer (TT) path and motion planning in narrow spaces.
In the general path and motion planning control architecture (see Fig. \ref{fig:overview}), we have identified the mid-level control (path following) as the bottleneck for navigating narrow complex environments with dynamic obstacles.
In particular, our work addresses two critical challenges for TT:
\begin{enumerate}
    \item computing real-time control inputs that produce dynamically feasible system trajectories even if the path planner does not identify such trajectories due to resolution and time constraints, and 
    \item active collision avoidance of dynamic obstacles within confined spaces under the assumption of an accurate motion prediction algorithm.
\end{enumerate}

These two challenges are fundamental for safe and reliable TT operation. 
We highlight that among our goals is to enable both forward and backward motion control of TT within a single framework which is directly applicable to real systems.
In particular, the TT reversing problem is especially challenging due to the unstable dynamics \cite{AltafiniSJ2002hscc,HejaseEtAl2018itsc}, i.e., small deviations in steering can lead to large, uncontrolled changes in the trailer angle and jackknifing.

In the remaining section, we present the motion model we consider in this paper, our assumptions, and the mathematical formulation of the control problem.


\subsection{Discrete‐Time Kinematic Model}

In this work, we focus in complex environments such as parking lots, unpaved camping grounds, loading docks, etc.
In such environments, the vehicles typically operate at low speed and thus a kinematic model of motion is sufficient, i.e., no-slip-angles assumption.
Even though we ignore tire forces, we still need to consider acceleration in the model in order to bridge the gap between the ideal simplified kinematic model and the high-fidelity black box model.
On the other hand, it is important to highlight that this assumption is not restrictive since 
(1) a kinematic model may be sufficient even for higher speed applications \cite{VallinderEtAl2024itsc}, and 
(2) MPPI based methods can handle any system model even purely data driven or hybrid dynamical \cite{parwana2025brmppi}.

\begin{assumption}
\label{ass:kin}
    We consider a kinematic model of motion which ignores tire forces, but uses jerk for input.
\end{assumption}

\begin{figure}[t]
    \centering
    \includegraphics[width=0.8\linewidth]{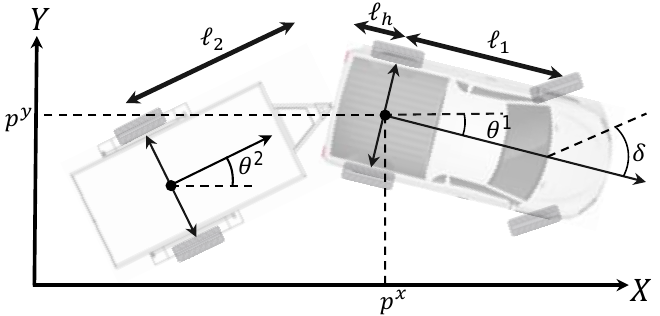}
    \caption{Tractor-trailer Kinematic Model.}
    \label{fig:tt_model}
\end{figure}

We employ Euler discretization with a sampling period \(\Ts\) to derive a discrete-time model of the TT system: 
\begin{equation}
\begin{aligned}
  \posX_{k+1}            &\!= \posX_k \! +\! \Ts\,\speed_k\,\cos\headingTractor_k\\
  \posY_{k+1}            &\!= \posY_k \! +\! \Ts\,\speed_k\,\sin\headingTractor_k\\
  \speed_{k+1}           &\!= \speed_k \! +\! \Ts\,\accel_k\\
  \accel_{k+1}           &\!= \accel_k \! +\! \Ts\,\jerk_k\\
  \headingTractor_{k+1}  &\!= \headingTractor_k \! +\! \Ts\,\frac{\speed_k}{\wheelbase}\,\tan\!\steerAngle_k\\
  \headingTrailer_{k+1}  &\!= \headingTrailer_k\!  \\
                        & \!+ \Ts\,\frac{\speed_k}{\trailerLen}
                             \Bigl(\sin(\headingTractor_k\!-\!\headingTrailer_k)
                             - \tfrac{\hitchLen}{\wheelbase}\cos(\headingTractor_k\!-\!\headingTrailer_k)\,\tan\!\steerAngle_k\Bigr)\\
  \steerAngle_{k+1}      &\!= \steerAngle_k \! +\! \Ts\,\steerRate_k
\end{aligned}
\label{eq:tt_discrete_kinematics}
\end{equation}
where $\posX_k,\posY_k$ are the tractor's rear‐axle center coordinates,
$\speed_k,\accel_k$ the longitudinal speed and acceleration, 
$\headingTractor_k,\headingTrailer_k$ the tractor and trailer headings, 
$\steerAngle_k$ the steering angle, 
$\jerk_k$ the longitudinal jerk, 
and $\steerRate_k$ the steering angle rate.  
The geometric parameters are the tractor wheelbase $\ell_1$, hitch length $\ell_h$, and trailer length $\ell_2$. 
The system parameters and positional variables are presented in Fig. \ref{fig:tt_model}.
Model  \eqref{eq:tt_discrete_kinematics} can be rewritten in the form of equation \eqref{eq:affine_dynamics} where 
$x_k\in\X$ is the state vector 
$
x_k = [\,\posX_k,\;\posY_k,\;\speed_k,\;\accel_k,\;\headingTractor_k,\;\headingTrailer_k,\;\steerAngle_k\,]^T,
$
and $u_k\in\U$ is the control input
$
u_k = [\,\jerk_k,\;\steerRate_k\,]^T.
$
Therefore, in the following, we will refer to \eqref{eq:tt_discrete_kinematics} as the system dynamics.


\subsection{Optimal Control Parking Problem}
We formulate the parking problem as a general Finite-horizon Optimal Control Problem (FHOCP) of finding a sequence of controls $\inpsig = (u_k)_{k=0}^{N-1}$ with $u_k\in\U$ 
that minimizes the cumulative running cost 
\begin{align}\label{eq:cost_function}
J\bigl(x,\inpsig\bigr)=
\sum_{k=0}^{N-1} \stageCost(x_k,u_k)+\stageCost_N(x_N),
\end{align}
given an initial state $x_0\in\X$ that steers the system \eqref{eq:affine_dynamics}
into a goal region $\X_g$ while staying within a time varying safe set $\X_k\subseteq\X$ for  $\{k\}^{N-1}_{k=0}$. 
Formally, the optimization problem is 
\begin{equation}
\begin{aligned}
\min_{\inpsig \in \U^{N}}\quad & 
  J\bigl(x,\inpsig\bigr)\\
\text{s.t.}\quad&
  \eqref{eq:affine_dynamics},~\text{for}~\{k\}^{N-1}_{k=0},\\
  &x_k \in \X_k,~ u_k\in \U,\quad \text{for}~\{k\}^{N-1}_{k=0},\\
  &x_N \in \X_g, ~ x_0 = x(0)\in \X_0.
\end{aligned}
\label{eq:general_parking_ocp}
\end{equation}
Here, $\stageCost:\X\times\U\to\mathbb{R}^+$ is the stage cost, $\stageCost_N:\X\to\mathbb{R}^+$ is the terminal cost, and $\X_g$ denotes the goal region (e.g., the set of states corresponding to the target parking spot). 
Note that optimization problem \eqref{eq:general_parking_ocp}
encodes any dynamic obstacles through the time varying safe set $\X_k$.

\begin{assumption}
\label{ass:prediction}
    In order to define $X_k$ up to the horizon $N-1$,     We assume that at every time $k$ we know the static obstacles and that for any dynamic obstacles, we have an accurate motion prediction algorithm. 
\end{assumption}

Assumption \ref{ass:prediction} can be lifted if we consider stochastic systems and/or environments along with state estimation algorithms, e.g., \cite{parwana2025brmppi,YangPML2023cdc}.
In the following, we present how to solve \eqref{eq:general_parking_ocp} with BR-MPPI under Assumptions \ref{ass:kin} and \ref{ass:prediction}.


\section{Motion Planning and Control via Barrier-rate Guided MPPI}
\label{sec:method}
This work employs the architecture shown in Fig.~\ref{fig:overview} to perform motion planning and control for TT parking maneuvers.
We assume that, prior to initiating the parking maneuver, the target parking spot is specified and a path planner is available to compute a feasible trajectory from the current pose of the TT system to the goal. 
Such planners typically operate on low-resolution maps and use simplified vehicle models to generate paths quickly, 
making them practical for real-time applications. 
In our experiments, we use a Hybrid A* planner to generate this reference path. 
However, our architecture is not limited to Hybrid A* and can integrate any planner capable of producing a feasible trajectory.

Since global planners do not account for real-time uncertainties, such as moving obstacles or tight spatial constraints, 
we employ Barrier-Rate guided MPPI (BR-MPPI) \cite{parwana2025brmppi} as a real-time tracking controller that refines the trajectory and ensures safe execution in dynamic environments with narrow spaces, 
as illustrated in Fig.~\ref{fig:overview}.
BR-MPPI extends the MPPI control framework by incorporating CBFs to enforce safety constraints during trajectory sampling. 
In our implementation, we further modify BR-MPPI to incorporate control bounds as additional safety constraints (see Sec.~\ref{sec:method_brmppi}).
At each control step, BR-MPPI receives the current state of the TT system and its environment, including static obstacles and predicted pedestrian motion, and generates control inputs by minimizing a tracking loss. 
Unlike standard MPPI methods that penalize collisions in the cost function, 
BR-MPPI projects sampled control inputs during rollouts onto a safe set defined by control barrier function (CBF) constraints, 
ensuring safety by construction (see Sec.\ref{sec:method_superellips_barrier}). 
To improve tracking performance, we propose using a contouring cost, which addresses issues common in trajectory tracking, such as sensitivity to path parameterization and poor alignment in sharp turns, by penalizing lateral and longitudinal deviations relative to the reference path (see Sec.\ref{sec:method_contouring_cost}). 
The optimized control inputs consist of longitudinal jerk and steering angle rate.
These inputs are passed to a low-level hitch angle stabilizer, implemented using a CBF-QP \cite{ames2016control, zeng2021safety} that ensures the hitch angle remains within a safe range (Sec.~\ref{sec:method_hitch_stabilizer}). 
The output of the stabilizer is then passed to a PID controller, which translates the jerk and steering angle rate into actuator commands: throttle, brake, and steering.
This control loop runs continuously until the TT system completes the parking maneuver. 
To fine-tune the performance of the entire control stack, we employ an automated hyperparameter tuning pipeline based on \cite{psy-taliro} that is a search-based test generation and falsification tool (Sec. ~\ref{section:experiments}).  
This process iteratively evaluates different sets of BR-MPPI and contouring cost parameters, using metrics such as tracking error, and clearance margin to identify configurations that best balance safety and tracking performance.

As shown in Fig.~\ref{fig:overview}, BR-MPPI rollouts explicitly account for dynamic (e.g., pedestrian) and static obstacles using super-ellipsoidal models, 
enabling the generation of safe and feasible trajectories in complex parking scenarios.
The following sections describe our modified BR-MPPI algorithm, the contouring cost formulation for path tracking, the barrier function models used for obstacle avoidance in BR-MPPI, and the hitch angle stabilization method.
\subsection{Barrier-Rate guided MPPI} \label{sec:method_brmppi}
Barrier-Rate guided MPPI (BR-MPPI) \cite{parwana2025brmppi} augments standard MPPI rollouts (Sec.~\ref{sec:prelim_mppi}) with explicit safety enforcement.  
Instead of trying to directly impose the inequality constraint \eqref{eq:barrier_constraint_prelim}, 
BR-MPPI imposes the following equality constraint for each constraint $i\in\{i\}^{N_o}_{i=1}$ at each prediction time $\tau$: 
\begin{align}\label{eq:cbf_equality}
    L_f\barrier_i(x_\tau) + L_g\barrier_i(x_\tau)\,u_\tau
= -\bigl(\alpha_{i,\tau} + u_{i,\tau}^{\alpha}\bigr)\,\barrier_i(x_\tau).
\end{align}
The defined equality constraint \eqref{eq:cbf_equality} allows the parameter $\alpha_i$ to change with time (hence the notation $\alpha_{i,\tau}\in \mathbb{R}$).
The variable $\alpha_{i,\tau}$ is initialized in the beginning of each rollout and is controlled by an auxiliary input $u_{i,\tau}^{\alpha}\in \mathbb{R}$. 
Here, $N_o$ specifies the number of safety constraints (e.g., the total number of surrounding dynamic and static obstacles). 
Given that an $\alpha_i$ variable is assigned for each constraint $i$, 
we define the vector of auxiliary inputs $u_{\tau}^{\alpha} = [u_{1,\tau}^{\alpha},\cdots,u_{N_o,\tau}^{\alpha}]^T$.
To promote safety while using the equality constraint \eqref{eq:cbf_equality}, BR-MPPI \cite{parwana2025brmppi} imposes a cost that motivates negative $\alpha_{i,\tau}$ values within a buffer zone around the obstacles to ensure $\dot{\barrier}_i(x) \geq 0$ near constraint boundaries. This approach maintains safety by projecting the sampling toward trajectories that satisfy the equality constraint \eqref{eq:cbf_equality} with safe margins.

At each rollout $s$, BR-MPPI samples the perturbed auxiliary inputs $\tilde u_{\tau}^{\alpha}$ and system inputs $\tilde u_{\tau}$.
The sampled inputs are then projected onto \eqref{eq:cbf_equality} by solving the weighted minimum-norm problem
\begin{subequations}\label{eq:projection}
\begin{align}
\min_{{u}_{\mathrm{proj}}, {u}_{\mathrm{proj}}^{\alpha}} \quad & \lVert {u}_{\mathrm{proj}} - \tilde{u}_\tau \rVert_{Q_1} + \lVert {u}_{\mathrm{proj}}^{\alpha} - \tilde{u}_\tau^{\alpha}\rVert_{Q_2} \\
\text{s.t.}\quad & L_f \barrier_i(x_\tau) + L_g \barrier_i(x_\tau)\,{u}_{\mathrm{proj}}\nonumber\\
&\quad= -(\alpha_{i,\tau}+{u}_{\mathrm{i,proj}}^{\alpha})\,\barrier_i(x_\tau), ~\mathrm{for}~ \{i\}^{N_o}_{i=1}\label{eq:projection_constraints},
\end{align}
\end{subequations}
where the obtained projected input ${u}_{\mathrm{proj}}$ gives the next predicted system state with \eqref{eq:affine_dynamics}, 
and ${u}_{\mathrm{proj}}^{\alpha}$ updates the next barrier rate $\alpha_{\tau+1}=\alpha_{\tau} + {u}_{\mathrm{proj}}^{\alpha}$.
Given that \eqref{eq:projection} is an equality-constrained weighted minimum-norm problem, 
an analytical solution exists for $u_{\mathrm{proj}}$ and $ u_{\mathrm{proj}}^{\alpha}$.  
However, it ignores actuator limits, so projected inputs may violate \(u\in[\underline u,\overline u]\) or \(\alpha\in[\underline\alpha,\overline\alpha]\).  
Hard bounds could be added as inequalities, but then no closed-form solution is available.  
Instead, letting $\augInput = [u^T_{\mathrm{proj}}, u^{\alpha^T}_{\mathrm{proj}}]^T\in[\augInputLow,\augInputHigh]$,
we incorporate the bounds as soft penalties and solve
\begin{subequations}\label{eq:softproj}
\begin{align}
\min_{\augInput}\quad
&\|\augInput - \augInputDes\|_{\Wmat}^2
+ \tfrac{\rho}{2}\Bigl(\|\augInput - \augInputLow\|^2 + \|\augInputHigh - \augInput\|^2\Bigr)\\
\text{s.t.}\quad
& A\,\augInput = b, \label{eq:projection_constraints_A_B}
\end{align}
\end{subequations}
where $\augInputDes=[\tilde u^T_{\tau},\tilde u^{\alpha^T}_{\tau}]^T$, $W=\mathrm{Diag}(Q_1,Q_2)$, $\rho\in\mathbb{R}^+$ is a penalty term,and \eqref{eq:projection_constraints_A_B} encodes \eqref{eq:projection_constraints} for all $i\in\{i\}^{N_o}_{i=1}$.  
A closed-form solution for \eqref{eq:softproj} is given by 
\begin{align} \label{eq:closed-form_projection}
    z^*=M^{-1} \Big[p-A^T(AM^{-1}A^T)^{-1}(AM^{-1}p -b)\Big],
\end{align}
where $M=W+\rho I$, $p=Wz_{\mathrm{des}}+\frac{\rho}{2} (\overline{z} + \underline{z})$, and $I$ is the identity matrix.
See \cite{parwana2025brmppi} for the derivation.
\subsection{Path Tracking via Contouring Cost} \label{sec:method_contouring_cost}

Given a feasible path from the Hybrid A* planner, we design the cost function ~\eqref{eq:cost_function} to minimize tracking error.  
In many real-time systems, the computed collision-free geometric path is typically resampled into a time-parameterized trajectory and then passed to a trajectory tracking controller. 
However, purely following a precomputed trajectory under model uncertainties or unknown disturbances (e.g. in reverse parking) can lead to error accumulation, which may cause sudden accelerations or even complete loss of tracking.
To address this, contouring approaches jointly minimize the approximated contouring error (the perpendicular distance between the TT’s current position and its projection onto the reference path) and the lag error (how far behind that projection lies).
This formulation provides a tunable way to trade off between tracking accuracy and forward progress toward the reference without any loss of tracking.
Inspired by works that integrate contouring within MPC \cite{kulic2024mpcc++,liniger2015optimization,lam2010model}, we propose the following cost for \eqref{eq:cost_function}:

    Assume that the reference path $\refPath(\contourProgress)$ is parameterized by arc length $\contourProgress\in[0,\refPathLen]$, 
so that any reference point is given by
$
\bigl(\refPosX(\contourProgress),\,\refPosY(\contourProgress),\,\refHeadingTractor(\contourProgress),\,\refHeadingTrailer(\contourProgress)\bigr),
$
where $\refPathLen$ is the total path length.  We obtain $\refPath(\contourProgress)$ by fitting a third‐order spline to the Hybrid A* output.  
The running cost $\stageCost(x_k,u_k,\contourProgress_k)$ in \eqref{eq:cost_function} is then defined as
\begin{align}
    r\big(x_k,u_k,&\contourProgress_k\big) = \lVert \contourErr(x_k,\contourProgress_k)\rVert^2_{\contourErrWeight} 
    + \lVert \lagErr(x_k,\contourProgress_k)\rVert^2_{\lagErrWeight} \nonumber\\
    &+\lVert \headingErr(x_k,\contourProgress_k)\rVert^2_{\headingErrWeight} 
    + \lVert u_k\rVert^2_{\controlEffortWeight} 
    +  \contourProgressRateWeight(\bar{\contourProgressRate} - \contourProgressRate_k)^2,
\label{eq:cost_function_contour}
\end{align}
where $\contourProgressRate_k=\dot\contourProgress_k$ is the projection velocity along the reference path. 
The final term in \eqref{eq:cost_function_contour} penalizes deviations from the desired progress rate $\bar{\contourProgressRate}$, 
which corresponds to the maximum possible progress. 
This term encourages the system to move steadily forward along the path while allowing temporary slowdowns for safety or accuracy.
The weight $\contourProgressRateWeight$ adjusts the trade-off between path advancement and tracking precision.
The control effort term $\lVert u_k \rVert^2_{\controlEffortWeight}$ regularizes the magnitude of the control inputs to discourage aggressive actions and improve smoothness.
The error terms $\contourErr(\cdot)$ and $\lagErr(\cdot)$ denote the contouring and lag errors, respectively, defined by
\begin{align}\label{eq:contour_lag_errors}
    \contourErr(x,u,\contourProgress) =&\sin{\big(\refHeadingTractor(\contourProgress)\big)}\Big(\posX-\refPosX(\contourProgress)\Big)\nonumber\\
    &\quad-\cos{\big(\refHeadingTractor(\contourProgress)\big)}\Big(\posY-\refPosY(\contourProgress)\Big),\\
    \lagErr(x,u,\contourProgress) =& -\cos{\big(\refHeadingTractor(\contourProgress)\big)}\Big(\posX-\refPosX(\contourProgress)\Big)\nonumber\\
    &\quad-\sin{\big(\refHeadingTractor(\contourProgress)\big)}\Big(\posY-\refPosY(\contourProgress)\Big).
\end{align}
The heading error
$
\headingErr(x,\contourProgress)
=[
  \headingTractor - \refHeadingTractor(\contourProgress),
  \headingTrailer - \refHeadingTrailer(\contourProgress)
]^T
$
measures the deviation of the tractor and trailer headings from their respective references.

\subsection{Super‐ellipsoidal Barrier Formulation for Obstacle Collision Avoidance}
\label{sec:method_superellips_barrier}

Each static obstacle (a surrounding parked vehicle) is approximated by a super‐ellipse centered at $(\obsPosX,\obsPosY)$ with semi‐axes $\obsSemiAxisX,\obsSemiAxisY$ and orientation~$\obsHeading$.  To account for the TT footprint, we cover the vehicle body with discs of radius $r$ and inflate each obstacle’s semi‐axes to $(\obsSemiAxisX + r,\;\obsSemiAxisY + r)$.  For each obstacle $o\in\Obst$, we define the barrier function
\begin{align}\label{eq:barrier_function}
\barrier_o(x)
=
\Bigl(\frac{\obsPosXPrime}{\obsSemiAxisX + r}\Bigr)^e
+
\Bigl(\frac{\obsPosYPrime}{\obsSemiAxisY + r}\Bigr)^e 
- 1,    
\end{align}
where $e\ge2$ is the super‐ellipse exponent, $\obsPosXPrime = \cos\obsHeading\,(\posX - \obsPosX) + \sin\obsHeading\,(\posY - \obsPosY)$, and $\obsPosYPrime = -\sin\obsHeading\,(\posX - \obsPosX) + \cos\obsHeading\,(\posY - \obsPosY)$.  
We encode the safe states as
$
\X_s \;=\;\bigcap_{o\in\Obst}\{\,x : \barrier_o(x)\ge0\}
$.
Since the system model \eqref{eq:tt_discrete_kinematics} has relative degree~3 with respect to the control inputs $u=[\jerk,\steerRate]$, we approximate each $\barrier_o$ over one sampling interval $\Ts$ via a second‐order Taylor expansion:
\begin{align}\label{eq:barrier_function_approx}
    \hat\barrier_o(x_k,u_k)
=
\barrier_o(x_k)
+ \Ts\,\dot{\barrier}_o(x_k)
+ \tfrac{\Ts^2}{2}\,\ddot{\barrier}_o(x_k),
\end{align}
where $\dot{\barrier}_o$ and $\ddot{\barrier}_o$ are the first and second Lie derivatives of $\barrier_o$ along the discrete‐time kinematics \eqref{eq:tt_discrete_kinematics}.  
For pedestrians, we use super-ellipsoids with shape exponent $e = 2$, resulting in circular level sets. 
In this case, we assume the pedestrian footprint is approximately circular, and set $\obsSemiAxisX = \obsSemiAxisY = r$, where $r$ is the pedestrian safety radius. 
As shown in Fig.~\ref{fig:overview}, pedestrians are visualized in yellow and surrounding parked vehicles in red, each represented using super-ellipsoidal regions with different shape exponents.

\subsection{Hitch Angle Stabilizer} \label{sec:method_hitch_stabilizer}

In reverse parking, small steering inputs can produce large articulations between tractor and trailer, so the hitch angle 
$
\hitchDelta = \headingTrailer - \headingTractor
$
may grow rapidly and risk jackknifing.  Moreover, the risk increases at higher speeds, so we introduce a velocity‐dependent threshold
$
\hitchThreshMod(x) \;=\; \hitchThresh \;-\; \velScale\,\speed,
$
where $\hitchThresh$ is the nominal maximum hitch angle and $\velScale>0$ scales the reduction as a function of the longitudinal speed $\speed_k$.
We then define the barrier
\begin{align}\label{eq:hitch_angle_barrier}
    \barrier_h(x) \;=\; \hitchThreshMod(x) \;-\; \bigl|\hitchDelta(x)\bigr|,
\end{align}
so that enforcing $\barrier_h(x_k)\ge0$ guarantees $|\hitchDelta|\le\hitchThreshMod$.  
At each time‐step $k$ we solve the CBF-QP
\begin{equation}
\begin{aligned}
u_k^* = \arg\min_{u\in\U}\quad & \|u - u_k^{\mathrm{ref}}\|_{Q_h}^2, \quad\text{s.t.}\quad\eqref{eq:barrier_constraint_prelim}
\end{aligned}
\label{eq:hitch_angle_qp}
\end{equation}
where $u_k^{\mathrm{ref}}$ is the BR-MPPI provided reference and $Q_h\succ0$.  

\section{Experimental Evaluation}
\label{section:experiments}
\begin{figure}[t]
    \centering
    \includegraphics[width=0.9\linewidth]{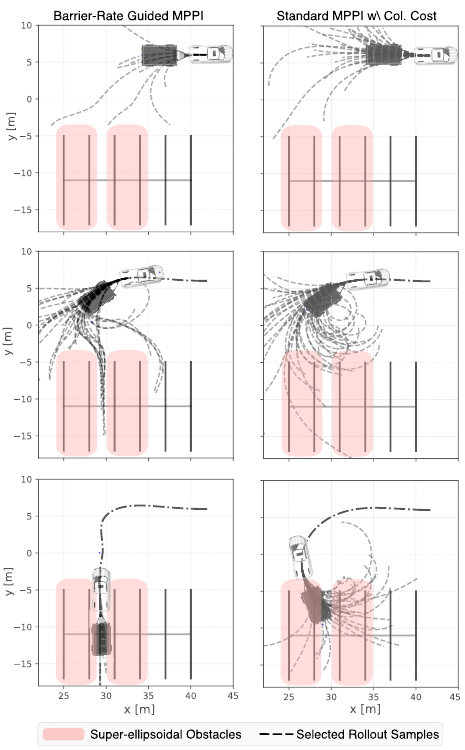}
    \caption{Rollout trajectories for backward parking using BR-MPPI and standard MPPI with collision cost.
    the figure shows generated rollouts at three different time frames. 
    Super-ellipsoidal obstacles represent inflated unsafe regions that account for the dimensions of the TT system. 
    BR-MPPI effectively projects rollouts outside these regions, enabling safe and stable backward maneuvers, 
    while standard MPPI struggles to avoid collisions.}
    \label{fig:vanilla_vs_BR}
\end{figure}
\begin{figure*}[t]
    \centering
    \includegraphics[width=0.9\linewidth]{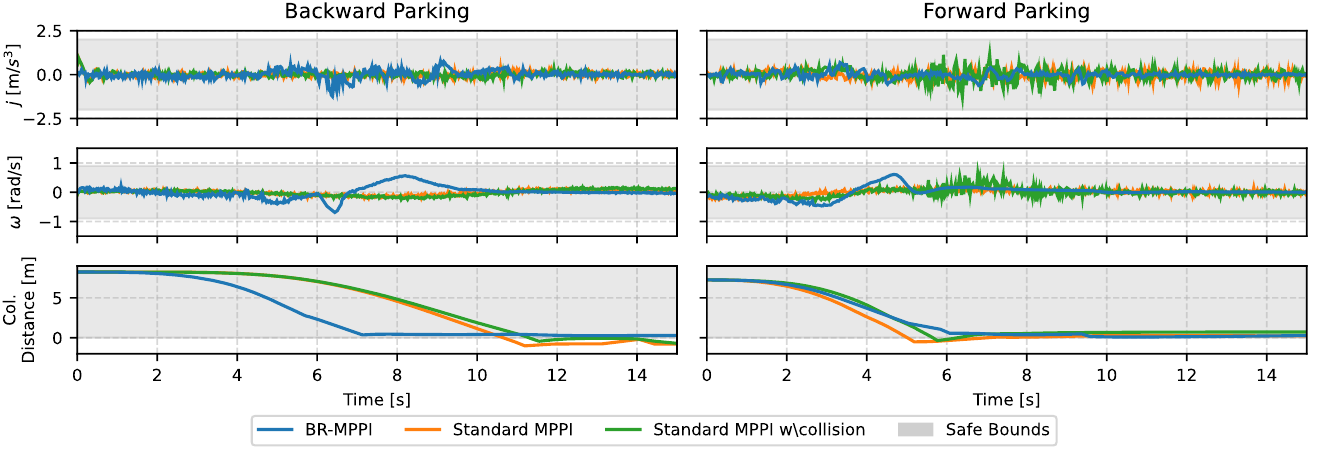}
    \caption{Control inputs and obstacle clearance signals during parking maneuvers.
    The plots show the control input signals and the minimum signed distance to the closest obstacle for backward parking (left) and forward parking (right). 
    Negative distance values indicate collisions. 
    BR-MPPI maintains safe distances while staying within control bounds in both scenarios.}
    \label{fig:signals}
\end{figure*}
To assess the effectiveness of BR-MPPI in narrow-space parking scenarios, we compare it against two standard MPPI baselines:
(1) MPPI is used solely to track a feasible path generated by a Hybrid A* planner,
(2) MPPI additionally incorporates a collision penalty while minimizing the tracking cost.
Obstacle distances are computed as the minimum distance between discretized bounding boxes around TT and the nearby obstacle. 
We adapted the existing Hybrid A* planner for the TT system from \cite{HybridAStarTrailer} and implemented our version in Python.
All MPPI variants, including BR-MPPI, are implemented in Python using the JAX library \cite{jax2018github} to enable GPU-accelerated rollouts. 
Our implementation parallelizes rollout generation and achieves execution times under 25 [ms] per MPPI step 
which is suitable for real-time use.
On average with eight obstacles, the controller achieves the computation time of 9.9 [ms] per MPPI step. 
For all methods, we use a prediction horizon of H = 120 steps (4.8 [s]) and generate 5000 rollouts per control step.

Experiments are conducted in a high-fidelity simulation environment using IPG CarMaker for both forward and backward parking scenarios.
Figure~\ref{fig:overview} illustrates the test scenario: the TT system is tasked with parking within a 3~m-wide designated green zone, located between two fully occupied adjacent parking spots. 
These parked vehicles are shown in red and are modeled as static obstacles using super-ellipsoids with exponent $e=4$ (as described in Sec.~\ref{sec:method_superellips_barrier}). 
Additionally, a dynamic pedestrian, shown in yellow, is modeled with a super-ellipsoid using exponent $e=2$ to represent a human crossing in front of the target parking spot just before the TT initiates its maneuver.
The nominal control trajectory $u_{\mathrm{nom}}$ is initialized with zero.

The tightest configuration tested that BR-MPPI successfully achieved collision-free parking involved initial distances of 6 [m] (x-axis) and 10 [m] (y-axis) between the target parking spot and the nearest axle of the TT system, i.e., the rear axle of the trailer in the backward and the front axle of the tractor in the forward cases. 

The TT vehicle model includes: tractor wheelbase $\ell_1 =$ 3.23 [m], trailer length $\ell_2 =$ 2.9 [m], hitch length $\ell_h =$ 1.15 [m], tractor width of 2 [m], and trailer width of 2.5 [m]. 
Control input bounds are set as $j \in [-2, 2]$ [m/s\textsuperscript{3}] (longitudinal jerk) and $\omega \in [-0.9, 0.9]$ [rad/s] (steering rate). 
We used \cite{psy-taliro} to tune the hyperparameters for all methods.

Figure~\ref{fig:vanilla_vs_BR} shows rollout samples generated by BR-MPPI and standard MPPI with collision cost in a scenario with only static obstacles.
Backward parking with a TT system is highly unstable due to jackknifing tendencies during acceleration. 
BR-MPPI effectively projects trajectories outside the unsafe regions, modeled as super-ellipsoidal obstacles, that lead to safe and stable maneuvers. 
In contrast, standard MPPI struggles to avoid collisions in tight spaces.

\begin{table}[t]
  \centering
  \caption{Parking results in forward and backward scenarios.
Tracking error denotes the mean deviation from the reference. Time indicates the average computation time per MPPI step. Clearance refers to the minimum signed distance to the closest obstacle (negative values indicate collision).}
  \label{tab:parking}
  \setlength\tabcolsep{5pt}
  \renewcommand{\arraystretch}{1.2}  
  \begin{tabular}{
      l l
      c  
      c  
      c  
  }
    \toprule
    & & \multicolumn{1}{c}{\makecell{Tracking\\Error [m]}} 
      & \multicolumn{1}{c}{\makecell{Time [ms]}} 
      & \multicolumn{1}{c}{\makecell{Clearance [m]}} \\
    \midrule
    \multirow{3}{*}{\textbf{Forward}} 
      & MPPI        & 0.43 &  3.20 & -1.05  \\
      & MPPI + Col.         & 0.44 & 13.00 & -0.75  \\
      & \cellcolor{gray!15}\textbf{BR-MPPI} 
                           & \cellcolor{gray!15}0.47 &  \cellcolor{gray!15}9.92 & \cellcolor{gray!15}\textbf{0.13} \\
    \midrule
    \multirow{3}{*}{\textbf{Backward}} 
      & MPPI        & 0.80 &  3.00 & -1.48  \\
      & MPPI + Col.         & 0.83 & 14.00 & -1.40  \\
      & \cellcolor{gray!15}\textbf{BR-MPPI} 
                           &\cellcolor{gray!15} \textbf{0.53} &\cellcolor{gray!15} 9.75 & \cellcolor{gray!15}\textbf{0.06} \\
    \bottomrule
  \end{tabular}
\end{table}

Table~\ref{tab:parking} reports the tracking error, MPPI computation time, and minimum signed distances to obstacles (negative values indicate collisions) for both forward and backward parking. 
BR-MPPI consistently avoids collisions, achieving an average clearance of 0.13 [m] (52\% of the available margin) in forward parking, and 0.06 [m] (24\%) in backward maneuvers. 
The standard baselines fail to park without crossing into the neighboring occupied spaces.
Despite the added barrier projection, BR-MPPI finishes each control update within 9.9 [ms] on average (below the 25 [ms] real-time budget). 
Conversely, the MPPI with collision-cost baseline increases runtime to 13 [ms]. 
In forward parking, all methods show comparable tracking performance, indicating that BR-MPPI's added safety does not come at the cost of accuracy.
In Backward parking, BR-MPPI results in lower tracking error that highlights its greater stability in reverse maneuvers.

Figure~\ref{fig:signals} compares control inputs and obstacle clearance over time. 
BR-MPPI respects control limits and maintains safe distances throughout the trajectory.
Overall, BR-MPPI delivers collision-free, precise parking in both directions while satisfying strict real-time constraints.

\section{Conclusion and Future Work}

We introduced a novel algorithm employing Control Barrier Function (CBF)-inspired conditions to direct trajectory sampling. Future work includes extending the approach to higher-order CBFs, specifically addressing cases where the control input appears beyond the first derivative of the barrier function, and establishing formal guarantees for inequality constraint satisfaction.
Additionally, we will conduct hardware evaluations across diverse systems to empirically validate the proposed framework. While not detailed here, our framework is anticipated to effectively address critical challenges in tractor-trailer (TT) systems. 
Specifically, BR-MPPI is expected to manage variable trailer loads and dynamic stability issues by incorporating learning models directly within the control loop, which will be the focus of subsequent research.


\bibliographystyle{IEEEtran}
\bibliography{ttcontrol.bib}

\end{document}